%% file: main.tex
\documentclass[conference]{IEEEtran}

\usepackage[
  style=ieee,
  minnames=1,
  maxnames=5,
  mincitenames=1,
  maxcitenames=1,
  hyperref=false,
  backref=false,
  dashed=true,
  ]{biblatex}
\usepackage[
  acronym,
  hyperfirst=false,
  nohypertypes={acronym}]{glossaries}
\usepackage[disable]{todonotes}

\usepackage{bm}
\usepackage{amsmath}
\usepackage{amsfonts}
\usepackage{stmaryrd}
\usepackage[tbtags]{mathtools}% lädt »amsmath« und verbessert es

\usepackage{subfig}
\usepackage{booktabs}
\usepackage{siunitx}
\usepackage{microtype}

\usepackage{url}
\usepackage{hyperref}
\usepackage[capitalise]{cleveref}

\newcommand{\citet}{\textcite}
\newcommand{\citep}{\parencite}

\ifCLASSINFOpdf
\else
\fi

\newcommand*{\mathvf}[1]{\ensuremath{\bm{#1}}}
\newcommand*{\vect}[1]{\mathvf{#1}}
\newcommand*{\tran}{^{\top}}

\newcommand*{\vecttran}[2][]{\ifthenelse{\isempty{#1}}{\vect{#2}\tran}{\vect{#2}_{#1}\tran}}

\newcommand{\mean}[1]{\overline{#1}}
\newcommand{\evalop}[1]{\left\llbracket #1 \right\rrbracket}
\newcommand{\normald}[1]{G_{\Omega}\left(\mathvf{x} = #1\right)}
\newcommand{\obs}[2]{\Psi_{#1}\left(#2\right)}
\newcommand{\obsd}[1]{\obs{\Omega}{#1}}
\newcommand{\normaldfunc}[2]{}
\newcommand{\mtv}{\mathvf{m}^{\left(t\right)}_{|V|}}
\newcommand{\obsmtv}{\obsd{ \mtv }}
\newcommand{\dmone}{d^{-1}}

\makeglossaries
\loadglsentries{sections/Acronyms.tex}

\begin{document}
\title{%
  Imitation Game: A Model-based and Imitation Learning Deep Reinforcement
  Learning Hybrid
}

\author{
  \begin{tabular}{ccccc}
    Eric MSP Veith\textsuperscript{1,2}
      & Torben Logemann\textsuperscript{1}
      & Aleksandr Berezin\textsuperscript{2}
      & Arlena Wellßow\textsuperscript{1,2}
      & Stephan Balduin\textsuperscript{2}
  \end{tabular}\\[0.75ex]
  \begin{tabular}{cc}
      \begin{minipage}{0.5\linewidth}\centering
        \textsuperscript{1}Carl von Ossietzky University Oldenburg\\
        Research Group Adversarial Resilience Learning\\
        Oldenburg, Germany\\
        Email: \texttt{firstname.lastname@uol.de}
      \end{minipage}
    &
      \begin{minipage}{0.5\linewidth}\centering
        \textsuperscript{2}OFFIS -- Institute for Information Technology\\
        R\&D Division Energy\\
        Oldenburg, Germany\\
        Email: \texttt{firstname.lastname@offis.de}
      \end{minipage}
  \end{tabular}
}
\maketitle

\begin{abstract}

  Autonomous and learning systems based on \acrlong{DRL} have firmly
  established themselves as a foundation for approaches to creating resilient
  and efficient \acrlongpl{CPES}. However, most current approaches suffer from
  two distinct problems: Modern model-free algorithms such as \acrlong{SAC}
  need a high number of samples to learn a meaningful policy, as well as a fallback to ward against concept drifts (e.\,g., catastrophic
  forgetting). In this paper, we present the work in progress towards a
  hybrid agent architecture that combines model-based \acrlong{DRL} with
  imitation learning to overcome both problems.

\end{abstract}


% \fi
\IEEEpeerreviewmaketitle

\input{sections/introduction}
\input{sections/related_work}
\input{sections/hybrid-discriminator-architecture}
\input{sections/case-study}
\input{sections/discussion}
\input{sections/future_work}

\section*{Acknowledgement}

This work was funded by the German Federal Ministry for Education and Research
(BMBF) under Grant No.~01IS22071.

The authors would explicitly like to thank Peter Palensky, Janos Sztipanovits,
and Sebastian Lehnhoff in their help in establishing the \gls{ARL} research
group.


\renewcommand*{\bibfont}{\small}
%\IEEEtriggeratref{36}
\printbibliography

\end{document}

%% file: sections/Acronyms.tex
\newacronym{ARL}{ARL}{Adversarial Resilience Learning}
\newacronym{AI}{AI}{Artificial Intelligence}
\newacronym{DQN}{DQN}{Deep Q Network}
\newacronym{ENV1}{Env-1}{Environment~1}
\newacronym{ENV2}{Env-2}{Environment~2}
\newacronym{DDPG}{DDPG}{Deep Deterministic Policy Gradient}
\newacronym{A2C}{A2C}{Advantage Actor Critic}
\newacronym{TRPO}{TRPO}{Trust Region Policy Optimization}
\newacronym{KL}{KL}{Kullback-Leibler}
\newacronym{PPO}{PPO}{Proximal Policy Gradient}
\newacronym{TD3}{TD3}{Twin-Delayed DDPG}
\newacronym{SAC}{SAC}{Soft Actor Critic}
\newacronym{SCADA}{SCADA}{Supervisory Control and Data Acquisition}
\newacronym{CPS}{CPS}{Cyber-Physical System}
\newacronym{ICT}{ICT}{Information and Communication Technology}
\newacronym{CHP}{CHP}{Combined Heat and Power}
\newacronym{CNI}{CNI}{Critical National Infrastructure}
\newacronym{MAS}{MAS}{Multi-Agent System}
\newacronym[
    longplural={Information Technologies}
]{IT}{IT}{Information Technology}
\newacronym{IoT}{IoT}{Internet of Things}
\newacronym{STIX}{STIX}{Structured Threat Information Expression}
\newacronym{TAXII}{TAXII}{Trusted Automated Exchange of Indicator Information}
\newacronym{CTI}{CTI}{Cyber Threat Intelligence}
\newacronym{MUC}{MUC}{Misuse-Case}
\newacronym{DoE}{DoE}{Design of Experiments}
\newacronym{MDP}{MDP}{Markov Decision Process}
\newacronym{POMDP}{POMDP}{Partially-Observable Markov Decision Process}
\newacronym{OOD}{OOD}{Out-Of-Distribution}
\newacronym{API}{API}{Application Programming Interface}
\newacronym{UML}{UML}{Unified Modelling Language}
\newacronym{XMI}{XMI}{XML Metadata Interchange}
\newacronym{VM}{VM}{Voltage Magnitude}
\newacronym{PDF}{PDF}{Probability Density Function}

\newacronym{MV}{MV}{Medium Voltage}
\newacronym{DSO}{DSO}{Distribution System Operator}
\newacronym{OT}{OT}{Operational Technology}
\newacronym{DER}{DER}{Distributed Energy Resource}
\newacronym{PV}{PV}{Photovoltaic}
\newacronym{VPP}{VPP}{Virtual Power Plant}
\newacronym{DT}{DT}{Decision Tree}
\newacronym{SMT}{SMT}{Satisfiability Modulo Theories}
\newacronym{EV}{EV}{Electric Vehicle}
\newacronym{CPES}{CPES}{Cyber-Physical Energy System}

\newacronym{RL}{RL}{Reinforcement Learning}
\newacronym{DRL}{DRL}{Deep Reinforcement Learning}
\newacronym{TVL}{TVL}{Ternary Vector List}
\newacronym{ANN}{ANN}{Artificial Neural Network}
\newacronym{DNN}{DNN}{Deep Neural Network}
\newacronym{GCNN}{GCNN}{Graph Convolutional Neural Network}
\newacronym{XRL}{XRL}{eXplainable Reinforcement Learning}
\newacronym{SGD}{SGD}{Stochastic Gradient Descent}
\newacronym{MCC}{MCC}{MountainCarContinuous-v0 environment}
\newacronym{PCA}{PCA}{Principle Component Analysis}
\newacronym{DL}{DL}{Deep Learning}
\newacronym{FF-DNN}{FF-DNN}{Feed-Forward \gls{DNN}}
\newacronym{RNG}{RNG}{Random Number Generator}
\newacronym{CMCL}{CMCL}{Curiosity Model}
\newacronym{MARL}{MARL}{Multi-Agent Reinforcement Learning}
\newacronym{SRL}{SRL}{State Representation Learning}
\newacronym{ML}{ML}{Machine Learning}

\newacronym{SHAP}{SHAP}{SHapley Additive exPlanations}

\newacronym{XAI}{XAI}{eXplainable Artificial Intelligence}
\newacronym{LP}{LP}{Linear Program}

\newacronym{IPCC}{IPCC}{Intergovernmental Panel on Climate Change}
\newacronym{UN}{UN}{United Nations}

%% file: sections/introduction.tex
\section{Introduction}
\label{sec:introduction}

Efficient operation of \gls{CNI} is a global, immediate need. The report of
the \gls{IPCC} of the \gls{UN} indicates that greenhouse gas emissions must
be halved by 2023 in order to constrain global warming to \SI{1.5}{\celsius}.
Approaches based on \gls{ML}, agent systems, and learning agents --- i.\,e.,
such based on \gls{DRL} --- have firmly established themselves in providing numerous aspects of efficiency increase as well as resilient
operation. From the hallmark paper that introduced \gls{DRL}
\parencite{mnih2013playing} to the development of MuZero
\parencite{Schrittwieser2019} and AlphaStar
\parencite{vinyals_grandmaster_2019}, learning agents research has inspired
many applications in the energy domain, including real power management
\parencite{veith2017universal}, reactive power management and voltage control
\parencite{Diao2019, thayer2020deep}, black start \parencite{wu_novel_2024},
anomaly detection \parencite{9956995}, or analysis of potential attack vectors
\parencite{veith2023learning,wolgast2023analyse}.

A promising, and also the simplest form to employ \gls{DRL}-based agents is
to make use of model-free algorithms, such as \gls{PPO}, \gls{TD3}, or
\gls{SAC}. Recent works show applications in, e.\,g., voltage control,
real power management in order to increase the share of \glspl{DER}, or
frequency regulation \parencite{chen2022reinforcement}.

However, model-free \gls{DRL} approaches suffer from prominent problems
that hinder their wide-scale rollout in power grids: low \emph{sample
efficiency} and the potential for \emph{catastrophic forgetting}. 

The first problem describes the way agents are trained. In \gls{DRL}, the
agent learns from interactions with its environment, receiving an indication
of the success of its actions through the \emph{reward} signal that serves as
the agent's utility function. Although modern model-free \gls{DRL} algorithms
show that they not only learn successful strategies but can also react very
well to situations they did not encounter during training, they require
several thousand steps to learn such a policy. As \gls{DRL} agents need
interaction with their environment (a simulated power grid in our case), 
making the training of \gls{DRL} agents computationally
expensive. Thus, the efficiency of one sample, i.\,e., the
quadruplet of state, action taken in the state, reward received, and follow-up
state, \((s, a, r, s')\), is low.

The second problem, catastrophic forgetting, describes that agents can act
suboptimal, erroneous even, or seem to “forget” established and well-working
strategies when introduced to a change in marginal distributions
\parencite{mccloskey_catastrophic_1989,khetarpal_towards_2020}. The
dominant literature concentrates on robotics and disjoint task sets; most
approaches cater to this specific notion of specific tasks. However, in the
power grid domain, there is no disjoint task set; instead, changes occur in the
environment, which can start catastrophic forgetting in a subtle way.

A particular way to address the first problem is the notion of model-based
\gls{DRL}. Here, the agent incorporates a model of the world, which helps to
not only gauge the effectiveness of an action but can also be queried internally
to learn from \parencite{wang_benchmarking_2019}. Potential models for such an
approach can also be surrogate models \parencite{hossain_efficient_2023}.
However, there currently is no architecture that also addresses the problem of
catastrophic forgetting.
Therefore, a clear research gap exists in constructing a \gls{DRL} agent approach that can make use of current advances
(i.\,e., specifically, train in an efficient manner), but is still reliable
for usage in power grids. Although it has been noted that a combination of
model-free \gls{DRL} and a controller would provide important benefits
\parencite{wang_integrating_2019}, a recent survey notes that currently, only
very limited work exists in this regard \parencite{chen2022reinforcement};
current approaches require a tight integration of the existing controller
logic with the \gls{DRL} algorithm \parencite{qu_combining_2020}. Aiding the
agent's learning by also adding imitation learning is currently not done.
However, current research seems to indicate that this is, in fact, necessary,
and that agents able to cope with various distributional shifts must have
learned a causal model of the data generator \parencite{richens2024robust}.

In this paper, we present such an agent approach. We describe the work in
progress towards a hybrid agent architecture that makes use of a world model
and can also efficiently learn from an existing policy, so-called
\emph{imitation learning}. Moreover, we use the existing, potentially
non-optimal control strategy as a safety fallback to guarantee a certain
behavior. We introduce the notion of a \emph{discriminator} that is able to
gauge between applying the \gls{DRL} policy and the fallback policy. We
showcase our approach with a case study in a voltage control setting.

The rest of this paper is structured as follows: In \cref{sec:related-work},
we give a survey of the relevant related work. We introduce our hybrid agent
and discriminator architecture in \cref{sec:hybrid-discriminator-architecture}.
We then describe our case study and obtained results in \cref{sec:case-study}.
We follow with a discussion of the results and implications in
\cref{sec:discussion}.  We conclude in \cref{sec:conclusion} and give an
outlook towards future results and development of this work in progress.

% vim:ft=tex:spell:spelllang=en

%% file: sections/related_work.tex
\section{Related Work}
\label{sec:related-work}

\subsection{Deep Reinforcement Learning}

%While model-based reinforcement learning is evolving fast (see the survey of
%\citet{luo2024survey}), this work is based on model-free reinforcement
%learning. A review regarding recent development in this field was presented by
%\citet{chen2022reinforcement}.

\Gls{DRL} is based on the \gls{MDP}, which is a quintuplet \((S, A, T, R, \gamma)\):
$S$ denotes the set of states (e.\,g., voltage magnitudes: 
\(S_t = \left\{ V_1(t), V_2(t), \dotsc, V_n(t) \right\}\)); $A$ is the set of
actions (e.\,g., the reactive power generation or consumption of a node the 
agent controls: \(A_t = \left\{ q_1(t), q_2(t), \dotsc, q_n(t) \right\}\));
$T$ are the conditional transition probabilities between any two states;
\(R\) is the reward function of the agent \(R: S \times A \rightarrow
\mathbb{R}\); and $\gamma$, the discount factor, which is a hyperparameter 
designating how much future rewards will be considered in calculating the absolute
\emph{Gain} of an episode.

Essentially, an agent observes a state \(s_t\) at the time \(t\), performs an action
\(a_t\), and receives a reward \(r_t\). Transition to the following state
\(s_{t+1}\) can be deterministic or probabilistic, depending on \(T\). 
The Markov property states that for each state \(s_t\) with an action \(a_t\), 
only the previous state \(s_{t-1}\) is relevant for the evaluation of the transition. 
To capture a multi-level context and maintain the Markov property, \(s_t\) is usually enriched with information from previous states or the relevant context.

The goal of reinforcement learning is generally to learn a policy in which
\(a_t \sim \pi_\theta(\cdot | s_t)\). The search for the optimal policy
\(\pi_\theta^*\) is the optimization problem on which all reinforcement learning algorithms are based.

\citet{mnih2013playing} proposed \gls{DRL} with their \glspl{DQN}.
The reinforcement learning itself precedes this publication
\citep{sutton2018reinforcement}. However, Mnih~et~al. were able to introduce deep neural networks as estimators for Q-values that enable robust training. 
Their end-to-end learning approach is still one of
the standard benchmarks in \gls{DRL}. The \gls{DQN} approach has seen
extensions until the Rainbow~DQN \citep{Hessel2018a} and newer work covers DQN approaches connected to behavior
cloning \parencite{li2022supervised}. Through, \Gls{DQN} only applies to environments with discrete actions.

\Gls{DDPG} \citep{Lillicrap2016} also builds on the policy gradient methodology: It concurrently learns a Q-function and policy. It is an off-policy algorithm that uses the Q-function estimator to train the policy.
\Gls{DDPG} allows for continuous control; it can be seen as \gls{DQN} for continuous action spaces. \Gls{DDPG} suffers from overestimating Q-values over
time; \gls{TD3} has been introduced to fix this behavior
\citep{fujimoto_addressing_2018}.

\Gls{PPO} \citep{schulman_proximal_2017} is an on-policy policy gradient algorithm that can be used for discrete and continuous action spaces. It is a development parallel to \gls{DDPG} and \gls{TD3}, not an immediate successor. \Gls{PPO} is more robust towards hyperparameter settings than
\gls{DDPG} and \gls{TD3} are. Still, as an on-policy algorithm, it requires more interaction with the environment train, making it unsuitable for computationally expensive simulations.

\Gls{SAC}, having been published close to concurrently with \gls{TD3}, targets
the exploration-exploitation dilemma by being based on entropy regularization
\citep{haarnoja_soft_2018}.  It is an off-policy algorithm originally focused on continuous action spaces but has been extended to support discrete action spaces as well. There also are approaches for distributed SAC
\cite{xie2021distributional}.

\Gls{PPO}, \gls{TD3}, and \gls{SAC} are the most commonly used model-free
\gls{DRL} algorithms today. Of these, off-policy learning algorithms are naturally suited for behavior cloning and imitation learning, as on-policy algorithms, by their nature, need data generated by the current policy to train.

Learning from existing data without interaction with an environment is called \emph{offline reinforcement learning}
\citep{prudencio_survey_2022}. 
This approach can be employed to learn from existing experiences. 
The core of reinforcement learning is the interaction with the environment. 
Only when the agent explores the environment, creating trajectories and 
receiving rewards, can it optimize its policy. However, more realistic environments, like robotics or the simulation of large power grids, 
are computationally expensive. Training from already existing data would be beneficial. For example, an agent could learn from an existing simulation run for optimal voltage control before being trained to tackle more complex scenarios.

The field of offline reinforcement learning can roughly be subdivided into \emph{policy constraints}, \emph{importance sampling}, \emph{regularization},
\emph{model-based offline reinforcement learning}, \emph{one-step learning},
\emph{imitation learning}, and \emph{trajectory optimization}. \citet{levine_offline_2020}\footnote{The tutorial by \citet{levine_offline_2020} is available only as a preprint. However, to our knowledge, it constitutes one of the best introductory seminal works so far. Since it is a tutorial/survey and not original research, we cite it despite its nature as a preprint and present it alongside the peer-reviewed publication by \citet{prudencio_survey_2022}, which cites the former, too.}
and \citet{prudencio_survey_2022} have published extensive tutorial and review papers, to which we refer the interested reader instead of providing a poor replication of their work here. Instead, we will give only a concise overview considering the 
ones relevant for this work.

Of the mentioned methods, imitation learning is especially of interest. It means the agent can observe the actions of another agent (e.g., an existing controller) and learn to imitate it. Imitation learning aims to reduce the 
distance between the policy out of the dataset \(D\) and the agent's policy, 
such that the optimization goal is expressed by \(J(\theta) =
D(\pi_\beta(\cdot | \bm{s}), \pi_\theta(\cdot | \bm{s}))\). This so-called 
Behavior Cloning requires an expert behavior policy, which can be hard to come by but is readily available in some power-grid-related use cases, such as voltage control, where a simple voltage controller could be queried as the expert.

\subsection{Model-free and Model-based Deep Reinforcement Learning for Power Grids}

As of today, a large corpus of works exists that consider the application of 
\gls{DRL} to power grid topics. As we have to constrain ourselves to works relevant to this paper, we refer the interested reader to current survey papers, such as the one created by
\textcite{chen2022reinforcement} for model-free and \citet{luo2024survey} for model-based learning, for a more complete overview.

One of the dominant topics for applying \gls{DRL} in power grids is voltage and reactive power control. As voltage control tasks dominate distribution systems, steady-state simulations can be employed, making the connection to \gls{DRL} frameworks comparatively easy and helping maintain the Markov property. Also, the design of the reward function is easy, allowing a concise notation such as \(R = -\sum_i (V_i - 1)\).

To capture the underlying physical properties of the power grid better, \textcite{lee_graph_2022} utilize \glspl{GCNN}. Other recent publications employ \gls{DDPG} for reactive power control, adding stability guarantees to the actor-network or training \cite{cao_multi-agent_2020, shi_stability_2022}. Model-based approaches can either incorporate offline learning from known (mis-) use cases \parencite{veith2023learning} or use surrogate models \parencite{hossain_efficient_2023}. \citet{gao2022model} use SAC and model augmenting for safe volt-VAR control in distribution grids.

Other applications include solutions to the optimal power flow problem
\parencite{woo_real-time_2020,zhou_data-driven_2020}, power dispatch
\parencite{lin_deep_2020,zhang_learning-based_2020}, or \gls{EV} charging
\parencite{silva_coordination_2020}. Management of real power also plays a
role in voltage control in distribution networks since conditions for
reactive-reactive power decoupling are no longer met due to comparable
magnitudes of line resistance and reactance.

% vim:ft=tex:spell:spelllang=en

%% file: sections/hybrid-discriminator-architecture.tex
\section{Hybrid Agent Architecture and Discriminator}
\label{sec:hybrid-discriminator-architecture}

Our approach incorporates two policies, a world model, as well as a
discriminator that chooses which policy to follow. It is part of the
\emph{Adversarial Resilience Learning} agent architecture
\parencite{Fischer2019,Veith2023}.

When an agent receives new sensor readings from the environment, usually, a
policy proposes the setpoints for actuators, which are subsequently applied.
In our case, we query two policies in parallel. The \emph{adaptive policy} is
based on \gls{SAC}, the deterministic \emph{rules policy} incorporates a
simple voltage controller. The voltage controller is based on the following
formula:

\begin{equation}\label{eq:var-control}
    \bm{q}(t+1) = \left[\bm{q}(t) - \bm{D}(\bm{V}(t)-1)\right]^+~,
\end{equation}

\noindent where the notation \([\cdot]^+\) denotes a projection of invalid
values to the range \([\underline{\bm{q}^g}, \overline{\bm{q}^g}]\), i.\,e.,
to the feasible range of setpoints for \(q(t+1)\) of each inverter. \(\bm{D}\)
is a diagonal matrix of step sizes.

As both policies return a proposal for setpoints to apply, the discriminator
has to choose the better approach. It does so by querying its world model. In
the simplest case, the world model is an actual model of the power grid.
However, we note that the power grid model will not capture dynamics such as
the behavior of other nodes or constraints stemming from grid codes (e.\,g.,
line loads or voltage gradients).

The quality of each decision proposal is quantified through the agent's
internal utility function (reward). Our reward function consists of three
elements,

\begin{enumerate}

  \item The world state as voltage levels of all buses:
    \(\mathvf{m}^{(t)}_{|V|}\);

  \item The buses observed by the agent, which is subset
    of the world state to account for partial observability:
    \(\Psi\left(\mathvf{m}^{(t)}_{|V|}\right)
    \subseteq \mathvf{m}^{(t)}_{|V|}\);

  \item the number of nodes controlled by agents that are still in service (i.\,e.,
    unaffected by grid code violations), \(\sum_b\evalop{ \obsmtv }_b\). Note
    that \(\llbracket \cdot \rrbracket\) are Iverson brackets.

\end{enumerate}

To express the importance of the voltage band of \(0.90 \ge V \ge
1.10\)\,\si{pu}, we map the voltage magnitude to a scalar by utilizing a
specifically shaped function borrowed from the Gaussian \gls{PDF}:

\begin{equation}
    G(\mathvf{x}, A, \mu, C, \sigma) =  \frac{A}{|\mathvf{x}|} \cdot \sum_x \exp \left(-\frac{\left(x - \mu\right)^2}{2\sigma^2} - C \right) 
\end{equation}
\begin{equation}
   G_\Omega(x) = G(\mathvf{x}, \mu = 1.0, \sigma = 0.032, C = 0.0, A = 1.0) 
\end{equation}

Weights can control each part's influence. We set \(\alpha = \beta = \gamma =
\frac{1}{3}\). We construct the agent's performance function to be:

\begin{multline}\label{eq:agent-performance}
%  \begin{multline}
    P_{\Omega}\left(\mathvf{m}^{(t)}\right)
    =\alpha \cdot \normald{\mathvf{m}^{(t)}_{|V|}}\\
    + \beta \cdot \mean{\normald{ \obsmtv }}\\
    +\gamma \cdot \left\{\sum_b{\evalop{ \obsmtv }_b}\right\}
    {\left\{\left| \mtv \right| \sum_b{\dmone}\right\}}^{-1}~.
%  \end{multline}
\end{multline}

Here, the term \(\overline{\bm{x}}\) denotes the mean of \(\bm{x}\).

The agent's performance function is normalized, i.\,e., \(0.0 \le
P(\vect{m}{(t)}{}) \le 1.0\) in general and specifically for
\(P_\Omega(\vect{m}{(t)}{})\).

The discriminator does not use the performance function value of each policy's
proposal directly. Instead, the performance values are tracked and averaged
over a time period \(t\) using a simple linear time-invariant function:

\begin{equation}
    \mathit{pt1}(y, u, t) = \left\{\begin{array}{l l}
        u & \quad \textrm{if \(t = 0 \)}\\
        y + \dfrac{u - y}{t} & \quad \textrm{otherwise,}
  \end{array} \right.
\end{equation}

This way, fluctuations, especially in the adaptive policy, will not lead to a
“flapping” behavior; instead, the discriminator will prefer the existing
controller for a longer period of time, switching to the adaptive policy only
when it provides beneficial setpoints throughout a number of iterations. The
dual-policy setup also treats the existing controller strategy as a fall-back,
should the performance of the \gls{SAC} policy drop (e.\,g., because of
catastrophic forgetting).

The discriminator effectively causes the adaptive policy to receive three
samples per step: Two projected (the proposal checked against the internal
world model) and the actual reward received by the environment. As \gls{SAC}
is an off-policy algorithm, it can learn from all three. Hence, we effectively
speed up the learning procedure.

\begin{figure}
    \centering
    \includegraphics[width=\linewidth]{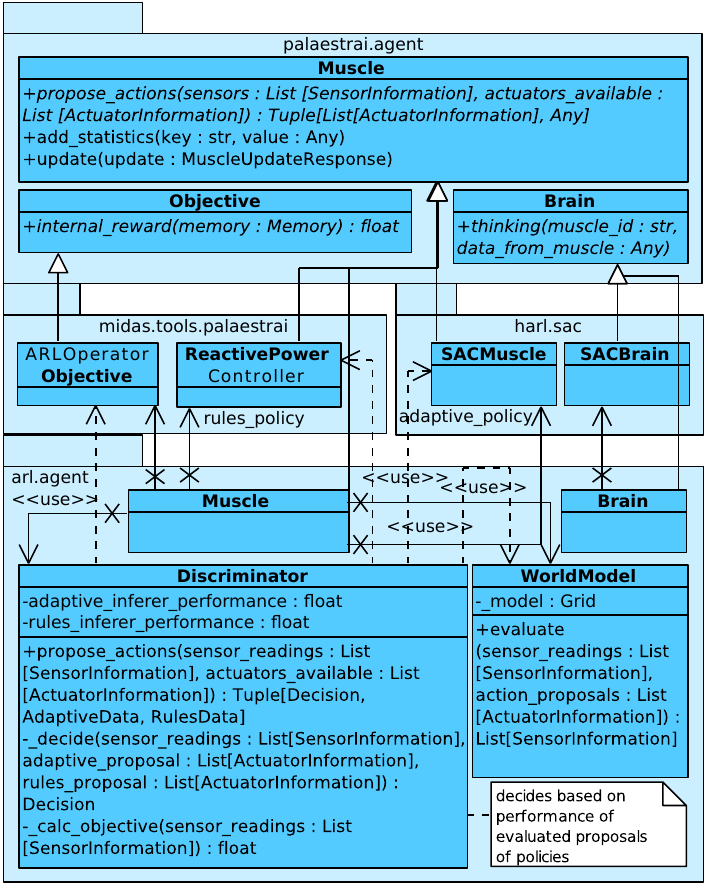}
    \caption{Class diagram of the ARL agent, featuring the Discriminator}
    \label{fig:discriminator_class_diagram}
\end{figure}

\Cref{fig:discriminator_class_diagram} depicts the architecture. Note that we
also include the \gls{DRL} rollout worker (\emph{Muscle}) and learner
(\emph{Brain}) classes to reference traditional \gls{DRL} architectures.

%% file: sections/case-study.tex
\section{Case Study}
\label{sec:case-study}

As an approach to validate this work-in-progress state, we have chosen the
CIGRÉ Medium Voltage benchmark grid. We have connected inverter-based loads
and generators to each busbar. The simulation does not include additional
actors, but it features constraints based on the German distribution system
grid code. 

The agent can control each load and generator, and is able to observe the
complete grid. Note that this actually poses a challenge for \gls{DRL}
algorithms: Loads consume reactive power, and generators can inject it. However,
an agent is able to use all actuators at once, even if this creates a
conflicting situation (generating VArs and consuming them at the same time).
Due to the possibility of grid code violations, the agent's internal world
model usually overestimates the reward generated by an action
(cf.~\cref{eq:agent-performance}).

\begin{figure*}
  \subfloat[SAC Agent]{\includegraphics[width=0.5\linewidth]{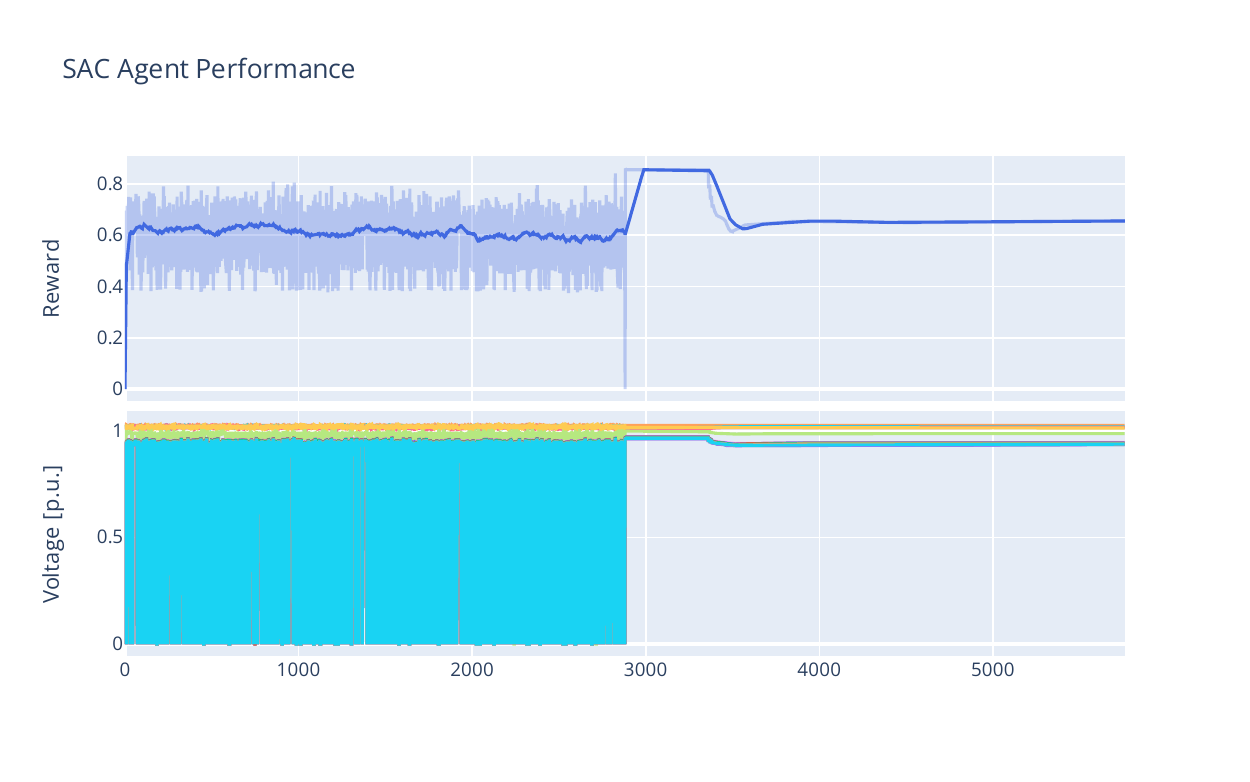}}~%
  \subfloat[ARL Agent]{\includegraphics[width=0.5\linewidth]{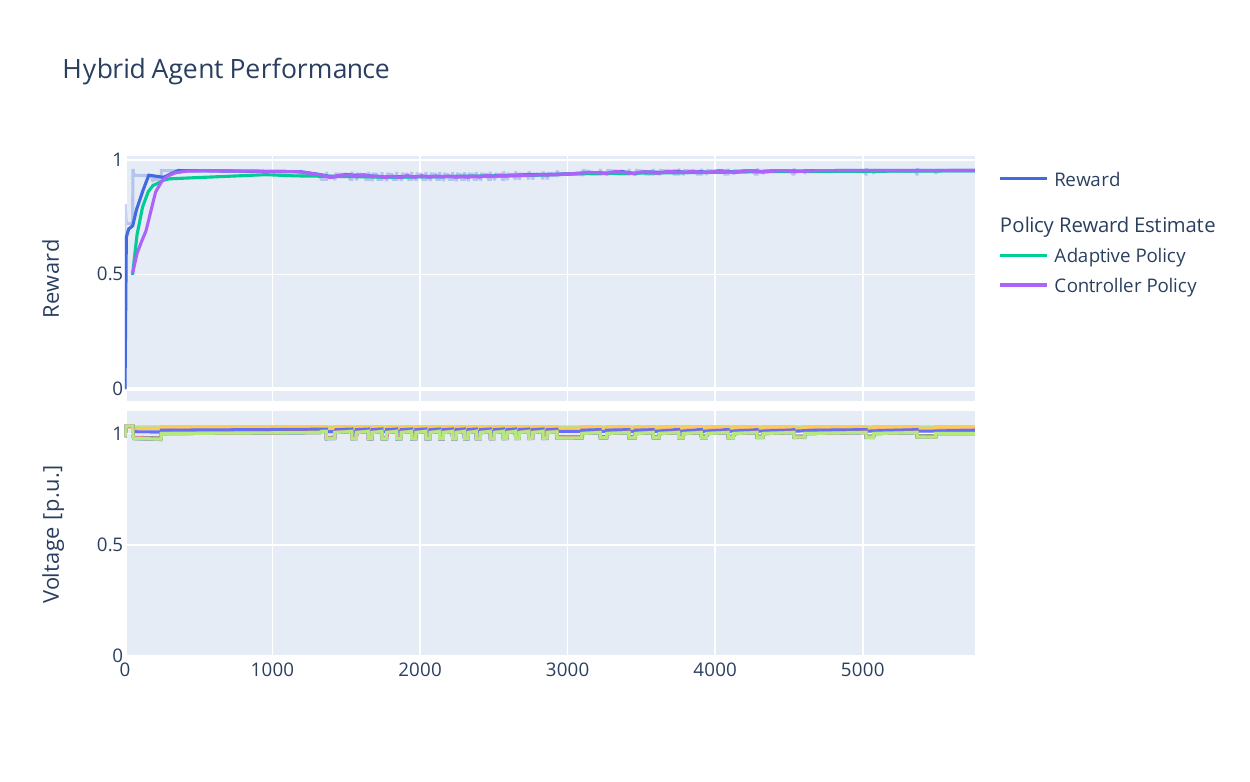}}
  \caption{Utility function results (performance) and voltage magnitudes by
    the pure SAC agent and the Hybrid approach}
  \label{fig:experiment-results}
\end{figure*}

We conduct training at test runs over \num{5760}~steps. We list all relevant
parameters to the grid and hyperparameters for the \gls{SAC} agents in
\cref{tab:parameters}. Results of the runs are depicted in
\cref{fig:experiment-results}. Here, we show a side-by-side comparison of the
same voltage control task, learned and executed by a pure \gls{SAC}-based
agent, and by our hybrid (ARL agent) approach. We show the agent's performance
in terms of its utility function, as per \cref{eq:agent-performance}. For the ARL
agent, we also note the policy estimates the discriminator makes based on the
world model.

\begin{table}
  \caption{Hyperparameters of SAC agents}
  \label{tab:parameters}
  \centering
  \begin{tabular}{cc}
    \toprule
    \textbf{Parameter} & \textbf{Value}\\
    \midrule
    Hidden Layer Dimensions & \((16, 16)\)\\
    Learning Rate & \(10^{-4}\)\\
    Warmup Steps & 50\\
    Training Frequency & every 5 steps\\
    \(\gamma\) & \num{0.9}\\
    \midrule
    Max. load per node & \SI{1.4}{MW}\\
    Max. reactive consumption p.\,n. & \SI{0.46}{MVar}\\
    Max. real power generation p.\,n. & \SI{0.8}{MW}\\
    Max. reactive power generation p.\,n. & \SI{0.46}{MVar}\\
    \bottomrule
  \end{tabular}
\end{table}

%% file: sections/discussion.tex
\section{Discussion}
\label{sec:discussion}

Based on \cref{fig:experiment-results}(a), one can discern the
training phase of the \gls{SAC} agent and the testing phase. The fluctuations
in the \gls{SAC} agent's reward curve are due to the violation of grid code
constraints, which happen because of the noise and entropy bonus \gls{SAC}
applies during training to explore the state/action space. These violations
are also visible in the voltage magnitudes plot, as buses disabled
due to grid code constraints have \SI{0}{p.u.} voltage magnitude.

In comparison, \cref{fig:experiment-results}(b) shows the results of the hybrid
(ARL) agent approach. There is no clear training/testing phase distinction
visible, which is because even during training, the fallback
option provided by the controller policy safeguards the agent against grid
code violations. For this reason, we have also plotted the internal reward
estimate of the discriminator. There is a gap due to the warm-up phase of the
internal \gls{SAC} learner; afterward, the \gls{SAC} agent learns quickly and
overtakes the controller policy.

We note that the hybrid agent approach does not cause any grid code
violations. Moreover, the overestimation of the grid code's state does not
yield any negative results. We assume that this is also due to the \gls{SAC}
algorithm's explicit remedy of overestimated Q values, which was the
Achilles' heel of \gls{DDPG}, and which now helps with the interaction of the
world model. In addition, the pure \gls{SAC} agent is not able to achieve the
theoretical maximum of the utility function. As we noted previously, the agent
has to learn the different actuators contradict each other, which the
\gls{SAC} implementation is not able to do during the approx. \num{5800} steps
it has to do so. In contrast, the imitation learning approach of our hybrid
(\gls{ARL}) agent does so, most probably not only due to a number of “correct”
samples provided by the controller, but also because it has thrice as many
samples in its replay buffer to train on, compared to the \gls{SAC} agent.

%% file: sections/future_work.tex
\section{Conclusions \& Future Work}
\label{sec:conclusion}

In this paper, we described a hybrid agent approach that incorporated both,
model-based \gls{DRL} as well as imitation learning into a single hybrid agent
architecture, which is part of the \gls{ARL} agent approach. We have provided
preliminary results that indicate that our approach leads to faster training
as well as guarantees on the benign behavior of the agent, which is able to
transparently alternate between a \gls{DRL} policy and a known and tried
controller policy.

In the future, we will extend our approach by testing it in more complex
scenarios, adding other actors, time series for \gls{DER} feed-in, more capable world models, as well as
adversarial agents. As the \gls{ARL} approach is a specific form of a
\gls{DRL} autocurriculum setup, we will evaluate the specific behavior of our
approach in the face of even misactors. We will also employ methods based on
\gls{DRL} \parencite{logemannNN2EQCDT2023} to estimate the effectiveness of
policies learned in this way. 
We expect that an autocurriculum approach is suitable especially to overcome 
adverse conditions, but has its own drawbacks in its nature of being a 
(model-free) \gls{DRL} approach. Here, we assume that our hybrid approach
will be able to bridge that gap and provide safety guarantees during 
normal operation, while being able to answer unexpected events by the virtue
of the \gls{DRL} policy.